# Predicting Stock Prices with FinBERT-LSTM: Integrating News Sentiment Analysis

WENJUN GU[1]

Carey Business School, Johns Hopkins University, Baltimore, USA

Yihao Zhong[2]

Courant Institute of Mathematical Sciences, New York University, New York, USA

Shizun Li[3]

Independent Researcher, Lynnwood, USA

Changsong Wei[4]

Digital Financial Information Technology Co.LTD, Chengdu, China

Liting Dong[5]

Cox School of Business, Southern Methodist University, Dallas, USA

Zhuoyue Wang[6]

Department of Electrical Engineering and Computer Sciences, University of California, Berkeley, Berkeley, USA

CHAO YAN*

Department of Electrical and Computer Engineering, Northeastern University, Boston, USA

yan.chao@northeastern.edu

The stock market's ascent typically mirrors the flourishing state of the economy, whereas its decline is often an indicator of an economic downturn. Therefore, for a long time, significant correlation elements for predicting trends in financial stock markets have been widely discussed, and people are becoming increasingly interested in the task of financial text mining. The inherent instability of stock prices makes them acutely responsive to fluctuations within the financial markets. In this article, we use deep learning networks, based on the history of stock prices and articles of financial, business, technical news that introduce market information to predict stock prices. We illustrate the enhancement of predictive precision by integrating weighted news categories into the forecasting model. We developed a pre-trained NLP model known as FinBERT, designed to discern the sentiments within financial texts. Subsequently, we advanced this model by incorporating the sophisticated Long Short Term Memory (LSTM) architecture, thus constructing the innovative FinBERT-LSTM model. This model utilizes news categories related to the stock market structure hierarchy, namely market, industry, and stock related news categories, combined with the stock market's stock price situation in the previous week for prediction. We selected NASDAQ-100 index stock data and trained the model on Benzinga news articles, and utilized Mean Absolute Error (MAE), Mean Absolute Percentage Error (MAPE), and Accuracy as the key metrics for the assessment and comparative analysis of the model's performance. The results indicate that FinBERT-LSTM performs the best, followed by LSTM, and DNN model ranks third in terms of effectiveness.

# 1 INTRODUCTION

The stock market is a key component of the economic ecosystem [1]. It serves as a conduit through which publicly traded corporations secure financial resources, which provides funding for various research and development projects to create services, products, and employment opportunities that contribute to economic growth. Should a company's performance falter, its share value is likely to experience a sharp decline; Should a company's performance excel, its stock value is likely to experience a dramatic surge. Investors are thus encouraged to delve into the intricacies of the stock market to discern which ventures are poised to yield a return. Anticipating stock price movements is inherently challenging, as it is not governed by a set of rigid mathematical formulas. The market is subject to fluctuation at any given moment, influenced by a myriad of elements including economic inflation and international tensions [2]. Nonetheless, by identifying trends within the data, it is possible to formulate reasonably precise short-term forecasts.

Analyzing current events can significantly contribute to forecasting stock movements, given that the stock market is profoundly swayed by news pertinent to the financial sphere [3]. News articles contain market information, negative articles are related to poor company performance, and positive articles are related to good performance. Therefore, it is possible to understand the trend of stocks by studying news articles, and incisive news analysis can yield substantial advantages by enhancing the accuracy of stock trend forecasts. Over recent years, individuals have examined stock-related news from a variety of angles, yet the potential for extracting valuable insights from financial news archives remains largely untapped. Despite the complexity introduced by a multitude of factors, the task of analyzing news remains a formidable challenge. Sentiment analysis, a subset of textual analysis, is employed to gauge the sentiment polarity of written content. This method evaluates the intrinsic sentiment of a text, categorizing it into positive or negative sentiments [4]. By doing so, it captures the public's emotional response to news stories. For instance, a report highlighting profits and acquisitions can evoke positive sentiments, potentially elevating a company's stock value, whereas a piece detailing layoffs and bankruptcy may incite negative reactions, contributing to a decrease in stock prices. Consequently, discerning the emotional undertones of news articles can offer a more profound insight into a company's operational success and inform predictions regarding its stock market trajectory.

Conversely, technical analysis techniques concentrate on examining the dynamics of stock prices, trading volumes, and the psychological expectations of investors. This approach leverages tools such as K-line charts to scrutinize the trajectory of stock indices for individual equities or the market as a whole, utilizing numerical data to forecast stock prices. Throughout history, a significant portion of initial scholarly work has focused on leveraging data from a particular moment, denoted as time t, to forecast the direction of stock prices at the immediately following moment, time t+1. In recent years, there has been a shift in the academic discourse, with some researchers framing the challenge of stock market prediction within the framework of sequential analysis. In this approach, the forecasting model processes a series of data points that span an ongoing timeline, as referenced in the literature [5-7].

# 2 STATE OF THE ART

Over an extended timeframe, the domain of predicting stock market movements has increasingly relied on the employment of sophisticated machine learning and deep learning methodologies, as highlighted in recent scholarly works [8-9]. Advancements in the field, notably the progression of Recurrent Neural Networks (RNNs), have been pivotal, especially the variants equipped with Long Short-Term Memory (LSTM) features, which have garnered significant attention in the literature [10], and the incorporation of attention mechanisms, notably self-attention and transformers [11], represent the cutting-edge advancements in deep learning. These sophisticated methods have notably enhanced the precision of tasks that are word-centric [13-14]. The predominant approach in academic literature for news-sensitive stock trend forecasting involves leveraging both stock price data and news articles. Moreover, technical indicators derived from stock price data are also commonly utilized. As highlighted in [15], this study brilliantly showcases the strength of ensemble methods in sentiment analysis, providing a robust framework for future research. Several studies have underscored the efficacy and relevance of integrating multifaceted data into predictive models [16-17]. Furthermore, it has been demonstrated that deep learning frameworks have significantly bolstered feature representation and forecasting accuracy within the financial sector [18-19]. In the work referenced as [20-22], an innovative dual-source stock market forecasting approach was presented, which utilizes numerical attention mechanisms, termed as Numerical Attention-Based (NBA). Subsequently, the study denoted by [23-25] introduced a state-of-the-art autoregressive neural network model that integrates elements of sentiment analysis into its predictive framework. They posit that the integration of predictive factors derived from news articles and Twitter activity can substantially elevate the precision of stock price forecasts. Building on this, [26-28] delves into the realm of stock price prediction through sentiment analysis of news, aiming to anticipate stock market movements.

This paper introduces a hybrid model that integrates sentiment analysis of financial news with stock price trend data, employing both Fin-BERT and Long Short-Term Memory (LSTM) networks. The structure of this paper is delineated as follows: Section 3 elaborates on the selection of pertinent methodologies, predominantly leveraging deep learning techniques, for dataset forecasting and stock market prediction. Section 4 presents and evaluates the efficacy of the Fin-BERT and LSTM models in concert with news headlines and stock price trends, juxtaposed with the performance of LSTM models that solely utilize stock price trends, as well as Deep Neural Network (DNN) models relying exclusively on stock price trends for their predictions. The work culminates in Section 5 with a summary of findings and suggestions for future research avenues.

## 3 METHODOLOGY

In this section, our work presents an in-depth account of the experimental procedures and evaluation metrics employed to validate the effectiveness of the proposed techniques and models. Our study leverages a comprehensive dataset of historical stock market news from Benzinga, complemented by US stock codes and pricing data sourced from Yahoo Finance. Utilizing the Keras open-source deep learning framework, which is integrated with a TensorFlow backend [10], we have constructed and refined convolutional neural network models. The execution of all experimental trials was conducted on cloud-based servers equipped with NVIDIA Tesla P100 GPU, which provides a unified platform using the NVIDIA Pascal GPU architecture.

### *3.1 Dataset*

The dataset used in this work is a carefully compiled collection of news information, containing a total of 843062 articles, published from February 15, 2009 to June 12, 2020. It comprehensively covers key information from multiple dimensions, aiming to provide a solid data foundation for in-depth insights into media trends, stock market dynamics, and their interrelationships. This dataset not only includes the title of each news article and the URL link to directly access the article content, ensuring the traceability of information and the possibility of instant access, but also accurately records the publisher information of each news article, which is particularly important for analyzing the reporting tendencies and influence distribution of different media. In addition, the dataset integrates the exact dates of news releases, spanning from February 15, 2009 to June 12, 2020, with a data coverage range of over eleven years, providing valuable time dimensions for time series analysis, trend evolution research, and tracing the influence of specific events on the stock market. Furthermore, the dataset is embedded with news related stock market information and corresponding stock codes, which is particularly prominent because it serves as a bridge for researchers to explore the immediate correlation and long-term effects between news events and stock market volatility. It has significant value in areas such as quantitative investment strategy development, market sentiment analysis, and risk management.

In addition, to ensure equitable representation in the training and validation datasets, it is imperative to employ a methodical strategy for data allocation. The original dataset was applied with simple hierarchical segmentation, with 85% of the data for each stock used in the training set and 15% used for testing. That is, a total of 716603 news information, stock prices, and other data were allocated to the training set, and 126459 news information, stock prices, and other data were designated to the test set. Also, in the training set data, 85% of the data for each stock was split for actual training, and 15% of the data was used for validation. That is, a total of 609113 news information, stock prices, and other data were assigned to the training set, and 107490 news information, stock prices, and other data were assigned to the validation set. as shown in table 1.

Table 1: Numbers of Daily Financial News and stock prices of dataset

| data | Daily Financial News and stock prices |
| --- | --- |
| Training | 609113 |
| validation | 107490 |
| testing | 126459 |
| Total no. of data | 843062 |

### 3.2 Fin-BERT Embedding LSTM Architecture

In this segment, we delve into the intricacies of the proposed model. The pipeline of this model first applies Fin-BERT word embedding to news data to generate news sentiment analysis indicators with ratings. At the same time, a LSTM is trained by combining historical stock price data to integrate the two models, enabling them to use all the features extracted from the two models (from "digital+news" data) to predict the closing price and reduce errors.

Fin-BERT is a language model based on BERT [23-24]. Fin-BERT represents an evolution in the domain of natural language processing (NLP), specifically tailored for financial sector applications. Its technological leap stems from the adoption of the Transformer architecture's bidirectional training mechanism for language modeling, marking a departure from prior studies [25] that predominantly focused on unidirectional, left-to-right text sequence analysis or a hybrid approach. Empirical evidence from scholarly work suggests that bidirectional training enables language models to achieve a more nuanced comprehension of linguistic contexts and flows. A key element of this architectural design is the integration of an attention mechanism, which is engineered to identify and understand the contextual interconnections between words and sub-word units within textual data. The BERT model stands out by featuring two principal components: an encoder that analyzes the input text and a decoder that generates task-specific predictions. Given BERT's objective to construct a comprehensive language model, it exclusively employs the encoder phase. This selective use of the encoder facilitates the model's ability to deduce word contexts from their immediate surroundings, thereby enhancing its interpretive capabilities. In our research, we harnessed the capabilities of the pre-trained Fin-BERT model, and its corresponding word segmentation to configure it for sequence classification tasks. The model will classify the input text into several predefined sentiment categories. They are neutral positive、negative。 We split the news headlines related to stocks into tokens and ensure that all input samples have the same length, with any gaps filled in with a special token "PAD". Employ the Softmax transformation to refine the initial logits produced by the model, thereby transforming them into a probabilistic distribution that aligns with the three distinct sentiment inclination percentages. This means that each sentiment category has a score between 0 and 1, and the sum of probabilities for all categories is 1. Calculate the average value of sentiment probability distribution groups. Then separate the tensor from the current computational graph and stop gradient tracking, which is necessary for inference as we do not need to calculate gradients to update model weights. Then, the calculated average emotional score is classified into the date of the news release, and statistical information including each date and its corresponding overall emotional tendency is obtained.

After obtaining overall sentiment statistics including each date and its corresponding stocks, we integrated emotional indicators (indifferent, favorable, unfavorable) with the closing prices from the preceding 8 trading sessions, transforming them into a format that is apt for input into a time series analytical model. The objective is to predict the closing price of the subsequent trading day, and prepared to input it into the LSTM model. We normalize the closing price data to map its value range to (0,1), and vertically concatenate the normalized closing price data with the previous closing price data. We employ a three-dimensional data structure, characterized by dimensions of (sample count, temporal sequence length, feature set size), to serve as the input for our analysis. The number of features includes 3 emotional features plus 1 feature for closing price data. The emotional data is copied to the first 3 feature positions of each time step, and the normalized closing price data is copied to the last feature position of the corresponding time step. We construct an input dataset that includes emotional features and historical closing price features of multiple time steps for each sample, and predict the closing price target for the next day corresponding to each sample. The architecture of the deep learning model is depicted in Figure 1, illustrating a sequential process where each stratum assimilates the preceding layer's output and subsequently forwards its own output to the subsequent stratum. The model will observe data from the past 8 time points as input to predict the output for the next time point. The first layer LSTM has 50 units and will return the output of the entire sequence, so that the next layer LSTM can continue to process this sequence information. The second and third layers of LSTM also have 50 units, which further extract patterns from the sequence. The second layer also returns the sequence output, while the third layer does not return the sequence. It aggregates information into a vector for final prediction. To conclude the model's construction, we incorporate a fully connected layer, commonly referred to as a Dense layer, designed to produce a single output dimension, thereby facilitating the generation of projected stock prices. The model employs a loss function based on the MSE metric, a prevalent choice for regression tasks that quantifies the discrepancy between the estimated and actual outcomes by averaging the squared deviations. Select the Adam optimizer for gradient descent and train at 100 epochs.

*3.3  LSTM Architecture*

In this method, unlike the data used in the previous section, we did not use the sentiment analysis score of stocks as input. Instead, we directly used the closing prices of stocks from the past 8 days as input data and reshaped the data into a two-dimensional array, where each row represents a sample containing the closing prices of 8 days, and each column corresponds to the closing prices of a time step. We normalized the data to a range of 0 to 1. Normalization accelerates the model's convergence in the training phase by mitigating the effects of scale disparities among various features. This process ensures the transformation of the initial time series data into a standardized form (mainly the closing prices of the past 8 days and a future day) into a format suitable for model input.

The configuration of the deep learning model demonstrate a cascade of layers where each one receives the preceding layer's result as its input and, in turn, conveys its own result to the following layer. The model will observe data from the past 8 time points as input to predict the output for the next time point. The first layer of LSTM has 50 neurons, which not only output predictions for the last time step, but also predictions for all time steps. The second layer of LSTM also has 50 neurons, continuing to return sequence outputs for all time steps and further extracting time series features. The third layer LSTM still has 50 neurons, but only outputs the result of the last time step. To culminate the model's architecture, a single-neuron fully connected layer is appended, tasked with yielding the ultimate forecast of the stock's closing value. The model opts for the MSE as its loss function, a metric that quantifies the model's predictive accuracy by averaging the squared deviations from the actual closing prices.

### 3.4 DNN Architecture

In this section, we also use the closing prices of stocks from the past 8 days as input data. The deep learning model's schematic illustrate a sequential arrangement of layers. Each layer integrates the output from its predecessor and subsequently forwards its own output to the next in line. The model initiates with a normalization layer to standardize the data, which is then processed through three fully connected layers containing 256, 128, and 64 neurons respectively. The constituent layers of the model employ a rectified linear unit (ReLU) activation function, augmented with a leaky parameter set to 0.01, addressing the issue of zero gradients for negative inputs. Concluding the sequence, a single-neuron fully connected layer is employed with a linear activation function, which is apt for regression analysis, and the model's objective is quantified by the MSE, a widely accepted metric for gauging the discrepancy between the predicted outcomes and the actual data points.

## 4 RESULTS

In this chapter, we conducted a detailed analysis and comparison of the predictive effects of three different models and methods on stock prices. Both the training and testing phases are conducted utilizing identical hardware setups. The training process encompasses a total of 100 epochs through the data. Additionally, a checkpoint mechanism is implemented to capture the optimal model parameters at various stages of training, with a focus on monitoring the model's performance based on the validation loss to ensure the most accurate predictions.

### 4.1 Performance Indicator

Performance metrics serve as critical tools for assessing the efficacy of machine learning and deep learning models. A variety of evaluative measures are at our disposal to gauge model performance. We have conducted a comparative analysis of the methodologies and models employed, focusing on three key metrics:

Mean Absolute Error: The Mean Absolute Error (MAE) represents the mean of the absolute differences between the actual target values and the model's predictions. The MAE is derived through the following formula:

$$MAE = \frac{\sum_{i=1}^{n}|y1-y2|}{n} \quad (1)$$

Where $y_1$ denotes the forecasted outcome for the $i^{th}$ training instance, and $y_2$ corresponds to the actual value for that instance.

Mean Absolute Percentage Error: The Mean Absolute Percentage Error (MAPE) serves as a measure that captures the average proportional discrepancy between forecasted and actual figures. It is derived by averaging the absolute percentage deviations, offering an insight into the model's accuracy in relation to the scale of the variable under consideration. MAPE can be calculated using the following formula:

$$MAPE = \frac{1}{n}\sum_{i=1}^{n}\left|\frac{y1-y2}{y2}\right| \quad (2)$$

Accuracy: Accuracy is obtained by subtracting MAPE from one.

$$Accuracy = 1 - MAPE \quad (3)$$

### 4.2 Fin-BERT Embedding LSTM Architecture

Our model achieved smaller loss within 77 epochs; Consequently, the model expedites the attainment of high precision. The validation dataset recorded a minimal loss value of 0.00036. The test dataset, as depicted in Figure 1, exhibited a slightly higher loss of 0.00083. Our model's predictions were characterized by a MAE of 173.67 and a MAPE of 0.045, alongside an impressive accuracy score of 0.955. To vividly illustrate the model's forecasting capabilities, we applied it to anticipate eBay's stock price trajectory for the subsequent 100-day period, as shown in Figure 2.

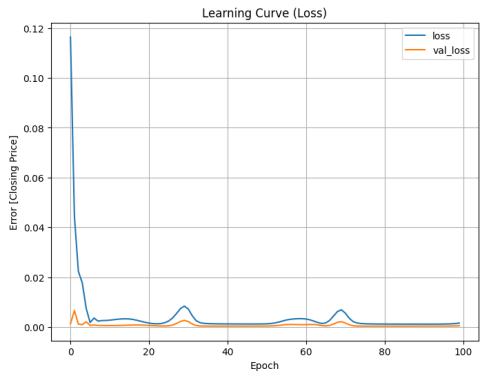
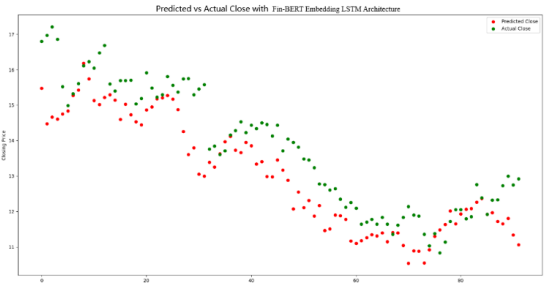

Fig. 1. Loss of Fin-BERT Embedding LSTM Architecture
Fig. 2. Predicted and Actual Close Price Compared under Fin-BERT Embedding LSTM Architecture

### 4.3 LSTM Architecture

Our model achieved smaller loss within 100 epochs. The validation set obtained a loss of 0.00085. The test dataset, as illustrated in Figure 3, reported a loss value of 0.00092. Our model, upon evaluation, produced a MAE of 183.36 and a MAPE of 0.072, achieving an accuracy level of 0.928. To provide a visual representation of the model's predictive prowess, we utilized it to forecast eBay's stock price movements for an upcoming 100-day period, as shown in Figure 4.

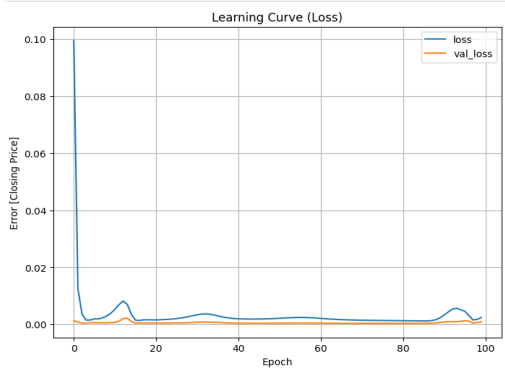 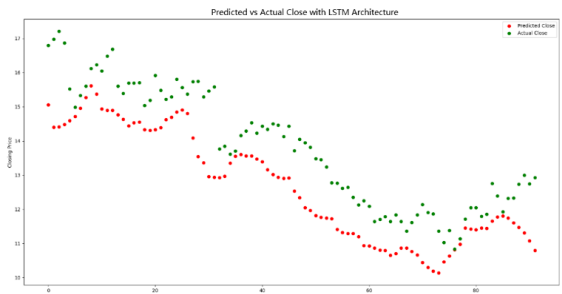

Fig. 3. Loss of LSTM Architecture

Fig. 4. Predicted and Actual Close Price Compared under LSTM Architecture

### 4.4 DNN Architecture

Our model achieved smaller loss within 100 epochs. The validation set obtained a loss of 0.458. The test set achieved a loss of 21.77, and our proposed model generated a MAE of 489.32 and a MAPE of 0.22, with an accuracy of 0.78. As shown in Figure 5. To visually demonstrate the predictive performance of the model, We deployed the model to project eBay's stock price trends for the forthcoming 100 trading days, as shown in Figure 6.

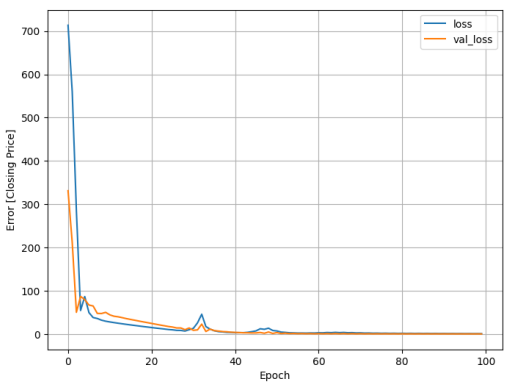 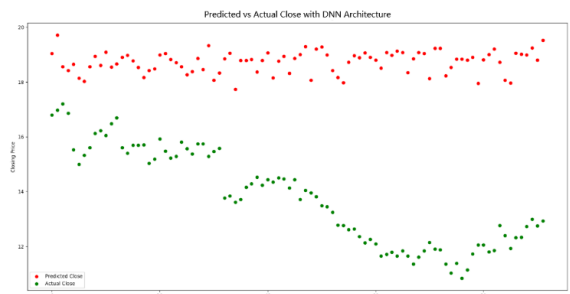

Fig. 5. Loss of DNN Architecture

Fig. 6. Predicted and Actual Close Price Compared under DNN Architecture

In conclusion, Fin-BERT Embedding LSTM Architecture and LSTM Architecture perform relatively well, while DNN Architecture performs the worst. In terms of testing loss, MAE, MAPE, and Accuracy, Fin-BERT Embedded LSTM Architecture performed the best, followed by LSTM Architecture, and DNN Architecture performed the worst. However, it is not difficult to see from the result data that the performance of Fin-BERT Embedding LSTM Architecture and LSTM Architecture is relatively close, while the performance of DNN Architecture is far from satisfactory. As shown in table 2.

Table 2: Comparison of different approach in terms of performance

| Approach | Testing loss | MAE | MAPE | Accuracy |
|---|---|---|---|---|
| Fin-BERT Embedding LSTM | 0.00083 | 173.67 | 0.045 | 0.955 |
| LSTM | 0.00092 | 183.36 | 0.072 | 0.928 |
| DNN | 21.77 | 489.32 | 0.22 | 0.78 |

## 5 CONCLUSION

Our work proposes three models and methods for predicting financial stock market prices, namely Fin-BERT Embedded LSTM Architecture, LSTM Architecture, and DNN Architecture. The Fin-BERT Embedding LSTM Architecture utilizes stock news content to quantitatively analyze news sentiment. At the same time, combining the past closing price trends of stocks with sequential data, the aim is to comprehensively extract features from the dual perspectives of numerical data and news information, to enhance the precision of forecasting stock closing prices and substantially diminish the predictive error. LSTM Architecture and DNN Architecture, on the other hand, only predict historical stock closing prices. The proposed Fin-BERT Embedding LSTM Architecture method performed the best, with a Testing loss of 0.00083, MAE of 173.67, MAPE of 0.045, and Accuracy of 0.955 at 77 epochs, resulting in better performance with less computation time. LSTM Architecture performed second, while DNN Architecture performed the worst. However, there is not much difference in performance between Fin-BERT Embedding LSTM Architecture and LSTM Architecture. In our future research, our goal will focus on two core directions: firstly, expanding the experimental scope, validating model performance on larger datasets, and deploying it to the current state-of-the-art model framework for in-depth testing. The second is to deepen the emotional analysis of stock market news, and it is expected that this strategy can significantly enhance the stability and reliability of model predictions. In addition, the plan is to include a fake news identification mechanism to distinguish the authenticity of financial reports, which is also expected to improve the robustness of predictions. The research perspective will not only be limited to the stock market, but also cross over to other investment sectors with huge potential, such as precious metals (such as gold), energy (oil), and the real estate industry, The aim is to validate and refine the model's performance across a variety of market conditions

**REFERENCES**

[1] Bosworth, B., Hymans, S., & Modigliani, F. (1975). The Stock Market and the Economy. Brookings Papers on Economic Activity, 1975(2), 257–300. https://doi.org/10.2307/2534104

[2] Liu, H., Shen, F., Qin, H., & Gao, F. (2024). Research on Flight Accidents Prediction based Back Propagation Neural Network. arXiv preprint arXiv:2406.13954.

[3] Jin, C., Huang, T., Zhang, Y., Pechenizkiy, M., Liu, S., Liu, S., & Chen, T. (2023). Visual prompting upgrades neural network sparsification: A data-model perspective. arXiv preprint arXiv:2312.01397.

[4] Jin, C., Che, T., Peng, H., Li, Y., & Pavone, M. (2024). Learning from teaching regularization: Generalizable correlations should be easy to imitate. arXiv preprint arXiv:2402.02769.

[5] Xie T, Wan Y, Wang H, Østrøm I, Wang S, He M, Deng R, Wu X, Grazian C, Kit C, Hoex B. Opinion Mining by Convolutional Neural Networks for Maximizing Discoverability of Nanomaterials. J Chem Inf Model. 2024 Apr 8;64(7):2746-2759. doi: 10.1021/acs.jcim.3c00746. Epub 2023 Nov 20. PMID: 37982753.

[6] Hu Z, Liu W, Bian J, Liu X, Liu T-Y, editors. Listening to chaotic whispers: A deep learning framework for news-oriented stock trend prediction. Proceedings of the eleventh ACM international conference on web search and data mining; 2018.

[7] Zhong, Y., Liu, Y., Gao, E., Wei, C., Wang, Z., & Yan, C. (2024). Deep Learning Solutions for Pneumonia Detection: Performance Comparison of Custom and Transfer Learning Models. *medRxiv*. https://doi.org/10.1101/2024.06.20.24309243

[8] Ni, H., Meng, S., Geng, X., Li, P., Li, Z., Chen, X., ... & Zhang, S. (2024). Time Series Modeling for Heart Rate Prediction: From ARIMA

to Transformers. arXiv preprint arXiv:2406.12199.

[9] Chong E, Han C, Park FC. Deep learning networks for stock market analysis and prediction: Methodology, data representations, and case studies. Expert Systems with Applications. 2017 Oct 15; 83:187-205.

[10] Xu, K., Wu, Y., Li, Z., Zhang, R., & Feng, Z. (2024). Investigating Financial Risk Behavior Prediction Using Deep Learning and Big Data. International Journal of Innovative Research in Engineering and Management, 11(3), 77-81.

[11] Xu, W., Chen, J., Ding, Z., & Wang, J. (2024). Text sentiment analysis and classification based on bidirectional Gated Recurrent Units (GRUs) model. arXiv preprint arXiv:2404.17123.

[12] Vaswani A, Shazeer N, Parmar N, Uszkoreit J, Jones L, Gomez AN, Kaiser Ł, Polosukhin I. Attention is all you need. Advances in neural information processing systems. 2017;30.

[13] Kalyani J, Bharathi P, Jyothi P. Stock trend prediction using news sentiment analysis. arXiv preprint arXiv:1607.01958. 2016 Jul 7.

[14] Thakkar A, Chaudhari K. Fusion in stock market prediction: A decade survey on the necessity, recent developments, and potential future directions. Information Fusion. 2021; 65:95–107. https://doi.org/10.1016/j.inffus.2020.08.019 PMID: 32868979

[15] Lin, Z., Wang, Z., Zhu, Y., Li, Z., & Qin, H. (2024). Text Sentiment Detection and Classification Based on Integrated Learning Algorithm. Applied Science and Engineering Journal for Advanced Research, 3(3), 27-33.

[16] Ni, H., Meng, S., Geng, X., Li, P., Li, Z., Chen, X., ... & Zhang, S. (2024). Time Series Modeling for Heart Rate Prediction: From ARIMA to Transformers. arXiv preprint arXiv:2406.12199.

[17] Liu, R., Xu, X., Shen, Y., Zhu, A., Yu, C., Chen, T., & Zhang, Y. (2024). Enhanced detection classification via clustering svm for various robot collaboration task. arXiv preprint arXiv:2405.03026.

[18] Shen, Y., Liu, H., Liu, X., Zhou, W., Zhou, C., & Chen, Y. (2024). Localization through particle filter powered neural network estimated monocular camera poses. arXiv preprint arXiv:2404.17685.

[19] Li X, Cao J, Pan Z. Market impact analysis via deep learned architectures. Neural Computing and Applications. 2019; 31(10):5989–6000.

[20] Ozbayoglu AM, Gudelek MU, Sezer OB. Deep learning for financial applications: A survey. Applied Soft Computing. 2020; 93:106384.

[21] Hu, X., Sun, Z., Nian, Y., Dang, Y., Li, F., Feng, J., ... & Tao, C. (2023). Explainable Graph Neural Network for Alzheimer's Disease And Related Dementias Risk Prediction. arXiv preprint arXiv:2309.06584.

[22] Ru, J., Yu, H., Liu, H., Liu, J., Zhang, X., & Xu, H. (2022). A bounded near-bottom cruise trajectory planning algorithm for underwater vehicles. Journal of Marine Science and Engineering, 11(1), 7.

[23] Li, X., Chang, J., Li, T., Fan, W., Ma, Y., & Ni, H. (2024). A Vehicle Classification Method Based on Machine Learning.

[24] Ni, H., Meng, S., Geng, X., Li, P., Li, Z., Chen, X., ... & Zhang, S. (2024). Time Series Modeling for Heart Rate Prediction: From ARIMA to Transformers. arXiv preprint arXiv:2406.12199.

[25] Devlin J, Chang MW, Lee K, Toutanova K. Bert: Pre-training of deep bidirectional transformers for language understanding. arXiv preprint arXiv:1810.04805. 2018 Oct 11.

[26] Yu, C., Xu, Y., Cao, J., Zhang, Y., Jin, Y., & Zhu, M. (2024). Credit card fraud detection using advanced transformer model. arXiv preprint arXiv:2406.03733.

[27] Zheng, Q., Yu, C., Cao, J., Xu, Y., Xing, Q., & Jin, Y. (2024). Advanced Payment Security System: XGBoost, CatBoost and SMOTE Integrated. arXiv preprint arXiv:2406.04658.

[28] Cao, J., Jiang, Y., Yu, C., Qin, F., & Jiang, Z. (2024). Rough set improved therapy-based metaverse assisting system. arXiv preprint arXiv:2406.04465.